\def\year{2022}\relax
\documentclass[letterpaper]{article} 
\usepackage{aaai22}  
\usepackage{times}  
\usepackage{helvet}  
\usepackage{courier}  
\usepackage[hyphens]{url}  
\usepackage{graphicx} 
\usepackage{natbib}  
\usepackage{caption} 
\DeclareCaptionStyle{ruled}{labelfont=normalfont,labelsep=colon,strut=off} 
\usepackage{rotating}
\usepackage{multirow}
\usepackage{amsmath}
\usepackage{amssymb}
\usepackage{booktabs}
\usepackage{algorithm2e}
\RestyleAlgo{ruled}

\title{Cross-Domain Few-Shot Graph Classification}
\author{Kaveh Hassani}
\affiliations{
    Autodesk AI Lab, Toronto, Canada \\
    kaveh.hassani@autodesk.com
}

\begin{document}

\maketitle

\begin{abstract}
We study the problem of few-shot graph classification across domains with nonequivalent feature spaces by introducing
three new cross-domain benchmarks constructed from publicly available datasets. We also propose an attention-based 
graph encoder that uses three congruent views of graphs, one contextual and two topological views, to learn representations 
of task-specific information for fast adaptation, and task-agnostic information for knowledge transfer. We run exhaustive 
experiments to evaluate the performance of contrastive and meta-learning strategies. We show that when coupled 
with metric-based meta-learning frameworks, the proposed encoder achieves the best average meta-test classification 
accuracy across all benchmarks.
\end{abstract}

\section{Introduction}
In Few-shot learning a model learns to adapt to novel categories from a few labeled samples. Common practices such as 
using augmentation, regularization, and pretraining may help in such a data-scarce regime, but cannot circumvent the problem. 
Inspired by human learning \cite{lake2015human}, meta-learning \cite{hospedales2020meta} leverages a distribution of similar tasks
\cite{garcia2018fewshot} to accumulate transferable knowledge from prior experience which then can serve as a strong inductive bias 
for fast adaptation to down-stream tasks \cite{sung2018learning}. In meta-learning, rapid learning occurs within a task whereas the 
knowledge about changes in task structure is gradually learned across tasks \cite{huang2020graph}. Examples of such learned knowledge 
are embedding functions \cite{vinyals2016matching, snell2017prototypical, garcia2018fewshot, sung2018learning}, initial parameters 
\cite{finn2017model, Raghu2020Rapid}, optimization strategies \cite{li2017meta}, or models that can directly map training samples to 
network weights \cite{garnelo2018conditional, mishra2018a}. 

A fundamental assumption in meta-learning is that tasks in meta-training and meta-testing phases are sampled from the 
same distribution, i.e., tasks are i.i.d. However, in many real-world applications, collecting tasks from the same distribution 
is infeasible. Instead, there are datasets available from the same modality but different domains. In transfer learning, this is 
referred as heterogeneous transfer learning where the feature/label spaces between the source and target domains are 
nonequivalent and are generally non-overlapping \cite{day2017survey}. It is observed that when there is a large shift between 
source and target domains, meta-learning algorithms are outperformed by pre-training/fine-tuning methods \cite{chen2018a}. 

A few work in computer vision addresses cross-domain few-shot learning by meta-learning the statistics of normalization 
layers \cite{Tseng2020cross,du2021metanorm}. These methods are limited to natural images that still contain a high degree 
of visual similarity \cite{guo2020broader}. Cross-domain learning is more crucial on variable-size order-invariant graph-structured data. 
Labeling graphs is more challenging compared to other common modalities because they usually represent concepts in specialized 
domains such as biology where labeling through wet-lab experiments is resource-intensive \citep{Hu2020Strategies} and labeling them 
procedurally using domain knowledge is costly \citep{Sun2020InfoGraph}. Furthermore, nonequivalent and non-overlapping feature
spaces is common across graph datasets in addition to shifts on marginal/conditional probability distributions. As an example, one may 
have access to small molecule datasets where each dataset uses a different set of features to represent the molecules \cite{day2017survey}. 

To the best of our knowledge, this is the first work pertaining cross-domain few-shot learning on graphs. To address this problem, we 
design a task-conditioned encoder that learns to attend to different representations of a task. Our contributions are as follows:   
\begin{itemize}
\item We introduce three benchmarks for cross-domain few-shot graph classification and perform exhaustive experiments to evaluate the 
performance of supervised, contrastive, and meta-learning strategies. 
\item We propose a graph encoder that learns to attend to three congruent views of graphs, one contextual and two topological views, to 
learn representations of task-specific information for fast adaptation, and task-agnostic information for knowledge transfer. 
\item We show that when coupled 
with metric-based meta-learning frameworks, the proposed encoder achieves the best average meta-testing classification accuracy 
across all three benchmarks.
\end{itemize}

\section{Related Work}
\textbf{Graph Neural Networks (GNNs)} are a class of deep models that combine the expressive power of graphs in modeling interactions with 
the unparalleled capacity of deep learning in learning representations. GNNs learn node representations over order-invariant 
and variable-size data, structured as graphs, through an iterative process of transferring, transforming, and aggregating the 
representations from topological neighbors. The learned representations are then summarized into a graph-level representation 
\cite{li_2015_iclr, gilmer_2017_icml, kipf_2017_iclr, velickovic_2018_iclr, xu_2019_iclr, Khasahmadi2020Memory, pmlr-v119-hassani20a}. 
GNNs are applied to non-Euclidean data such as point clouds \cite{ hassani_2019_iccv}, robot designs \cite{wang_2018_iclr}, physical 
processes \cite{pmlr-v119-sanchez-gonzalez20a}, molecules \cite{duvenaud_2015_nips}, social networks \cite{kipf_2017_iclr}, and knowledge 
graphs \cite{vivona_2019_nips}. For an overview on GNNs see \cite{zhang_2020_kde}.

\noindent \textbf{Meta-Learning on Graphs} is an under-addressed problem. A few works including Meta-GNN \cite{zhou2019meta}, G-Meta 
\cite{huang2020graph}, and GFL \cite{yao2020graph} focus on few-shot node classification via meta gradients, whereas other 
work such as MetaR \cite{chen2019meta}, GMatching \cite{xiong2018one}, and Meta-Graph \cite{bose2019meta} focus on few-shot link prediction 
and generalization over relations in knowledge graphs. More relevant to this study, MetaSpecGraph \cite{Chauhan2020Few} uses a super-class 
prototypical network for few-shot graph classification where super-classes are constructed by clustering graphs using spectrum of normalized 
Laplacian. All these works assume i.i.d distribution across the tasks which makes them infeasible for real-world applications. On the other hand, 
we for the first time address the problem of few-shot graph classification within the cross-domain setting.

\noindent \textbf{Domain Adaptation} is a type of transductive transfer learning in which the source and target 
classes are equivalent but the domains are different, and the goal is to reduce the distribution shift between the 
two domains \cite{zhuang2020comprehensive, wilson2020survey}. Most work addresses it by reducing the shift in input, feature, or output spaces using adversarial training \cite{tzeng2017adversarial, chen2018domain, 
hoffman2018cycada}. A major drawback of these methods is that they require access to unlabeled 
samples from the target domain during the training which makes them less practical \cite{Tseng2020cross}. 
For a review see \cite{wilson2020survey}. 

\noindent \textbf{Domain Generalization} aims to generalize from a set of seen domains to unseen domains without 
knowledge about the target distribution during training \cite{dou2019domain,Tseng2020cross}. Different 
strategies such as adversarial data augmentation \cite{volpi2018generalizing, shankar2018generalizing}, 
extracting task-specific domain-invariant features \cite{li2018deep, akuzawa2019domain}, or learning-to-learn 
strategies to simulate the generalization process \cite{dou2019domain,li2018learning, li2019feature} are used to tackle this problem. Domain generalization is more challenging in few-shot setting. On proper 
\emph{cross-domain few-shot} benchmarks, meta-learning methods are outperformed by simple fine-tuning 
and even in some cases by networks with random weights \cite{guo2020broader}. Only a few methods are 
introduced for cross-domain few-shot learning. Feature-wise transform (FWT) \cite{Tseng2020cross} uses 
feature-wise transform layers to encourage learning representations with an improved ability to generalize 
whereas MetaNorm \cite{du2021metanorm} learns to infer adaptive statistics for batch normalization. We use a 
combination of learning task-specific domain-invariant features and inductive normalization layers to achieve 
domain generalization in cross-domain few-shot graph classification setting. 

\section{Problem Formulation}
A \emph{domain} $\mathcal{D}=\{\mathcal{X}, \mathcal{Y}, P_{\mathcal{X}, \mathcal{Y}}\}$ is defined as a joint distribution $P_{\mathcal{X}, \mathcal{Y}}$ over the feature space $\mathcal{X}$ and label 
space $\mathcal{Y}$. We denote the marginal distribution over feature space as $P_{\mathcal{X}}$ and a parametric model over joint 
distribution as $f_\theta: \mathcal{X}\longmapsto \mathcal{Y}$ where $f_\theta(x)=\{P(y_k|x, \theta)\ |\ y_k \in \mathcal{Y}\}$. The model parameters are learned by 
minimizing the expected error over loss function $\mathcal{L}$: $\mathbb{E}_{(x,y) \sim P_{\mathcal{X}, \mathcal{Y}}}\left[\mathcal{L}(f_\theta(x), y)\right]$. In cross-domain few-shot 
learning, it is assumed that two domains exist: source domain $\mathcal{D}_S$ and target domain $\mathcal{D}_T$ such that their 
marginal distributions are different $P_{\mathcal{X}_S} \neq P_{\mathcal{X}_T}$, and $\mathcal{Y}_S$ and $\mathcal{Y}_T$ are disjoint. The source domain is available 
during meta-training phase whereas the target domain is only seen in meta-testing phase. During 
meta-training, tasks $\{\mathcal{T}_i | i=1...N \}$ are drawn from a distribution of tasks defined over $\mathcal{D}_S$, i.e., $\mathcal{T}_i \sim P_S(\mathcal{T})$, 
where each task consists of two non-overlapping small datasets:$D_i^{support}=\{(x_j, y_j)\}_{j=1}^{k\times n}, D_i^{query}=\{(x_j, y_j)\}_{j=1}^{k\times m}$. $k$ denotes the number of sampled classes and $n$ and $m$ are 
number of examples per category, i.e., \emph{k-way n-shot learning}. 

In the meta-training phase, the model's error on the support set provides \emph{task-level} update signals, 
while error on the query set after the model adapts to the support set, provides \emph{meta-level} update 
signals. During the meta-testing stage, the model is expected to quickly adapt to task $\mathcal{T}_j \sim P_T(\mathcal{T})$ by only 
accessing the support set for that task. The tasks in the meta-testing phase are sampled from $\mathcal{D}_T$ and 
$P_S(\mathcal{T}) \neq  P_T(\mathcal{T})$. It is noteworthy that learning in cross-domain few-shot setting is more difficult than 
learning in the transductive setting of traditional meta-learning where $\mathcal{D}_S =\mathcal{D}_T$ and as a result $ P_{\mathcal{X}_S} = P_{\mathcal{X}_T}$ and $P_S(\mathcal{T}) =  P_T(\mathcal{T})$.

In cross-domain few-shot learning, it is assumed that: (1) there is a domain shift between the source and target 
domains ($ P_{\mathcal{X}_S} \neq P_{\mathcal{X}_T}$), and (2) the feature spaces are equivalent between domains ($\mathcal{X}_S = \mathcal{X}_T$). This makes 
sense in computer vision where various image acquisition methods such as satellite images, dermatology 
images, and radiology images share a similar feature space with natural images \cite{guo2020broader}. 
However, in domains where data is represented as graphs, this assumption does not hold. For example, two 
molecular property prediction datasets may have different node/edge feature spaces due to the method used 
to generate the datasets (e.g., one may contain additional atom features such as formal charge and whether the 
atom is in the ring) \cite{hu2020open}. 

As such, we go beyond cross-domain few-shot learning and investigate its heterogeneous variant for graph 
classification. In this setting, we assume that: (1) Each task is possibly sampled from a dedicated domain 
different from all other tasks, either in meta-training or meta-testing phase, i.e. if there are $N$ tasks in 
meta-training and $M$ tasks in meta-testing phases, there exist $|\mathcal{D}|\le N+M$  domains. (2) Tasks are heterogeneous, which means they may have nonequivalent non-overlapping feature spaces, i.e., different 
dimensions in addition to distribution differences and disjoint label spaces:
$\forall i \neq j: \mathcal{X}_{\mathcal{T}_i} \neq \mathcal{X}_{\mathcal{T}_j},  P_{\mathcal{X}_{\mathcal{T}_i}} \neq P_{\mathcal{X}_{\mathcal{T}_j}}, \mathcal{Y}_{\mathcal{T}_i} \neq \mathcal{Y}_{\mathcal{T}_j}$. (3) Tasks can be grouped based on their \emph{meta-domains} 
which essentially defines the conceptual domain of a task. For example, all datasets that represent social 
networks can be grouped under a \emph{social network} meta-domain despite the fact that they may 
have different feature and label spaces. We investigate whether there exists underlying knowledge that can be transferred across these meta-domains.

\section{Method}
\begin{figure*}[ht] 
\label{figarch}
\begin{center}
\centerline{\includegraphics[width=175mm]{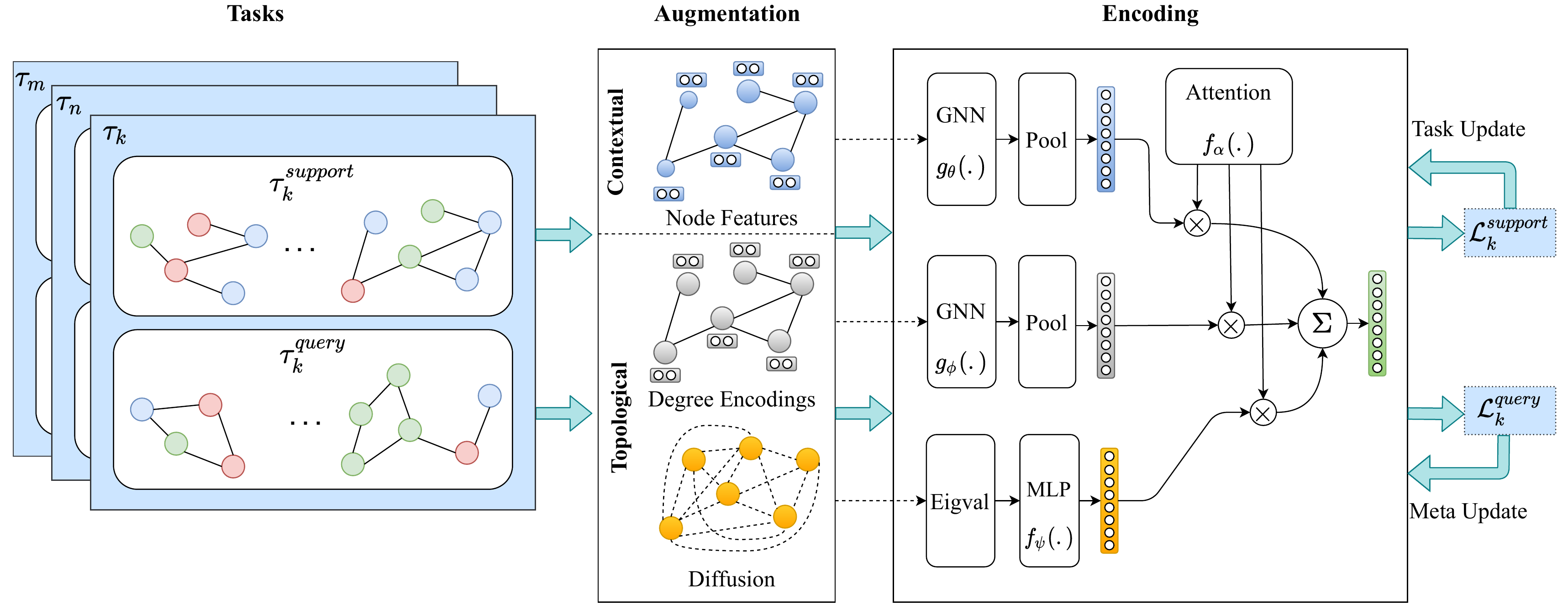}}

\caption{The proposed model for cross-domain few-shot graph classification. Graphs from sampled tasks are 
augmented to one contextual view and two geometric views and fed to three dedicated encoders resulting in 
three representations of the same graph. An attention mechanism is then used to aggregate the representations 
into a single graph representation. The parameters of the encoders and the attention mechanism are learned end-to-end using an arbitrary metric-based meta-learning approach.}
\end{center}
\end{figure*}

Graph-structured data can be analyzed from two congruent views: a contextual view and a topological view. 
The contextual view is based on initial node or edge features (for simplicity and without loss of generality, we 
only consider node features) and carries task-specific information. The topological view, on the other hand, 
represents topological properties of a graph which are task-agnostic and hence can be used as an anchor to align graphs from various domains in the feature space. We exploit this dual representation and explicitly 
disentangle them by designing dedicated encoders for each view which in return imposes the needed 
inductive bias to learn task-specific domain-invariant features. In a heterogeneous few-shot setting, the 
topological features can help with knowledge transfer across tasks whereas the contextual features can help 
with fast adaptation. We also use an attention mechanism that is implicitly conditioned on the tasks and learns 
to aggregate the learned features from the two views. We use a meta-learning strategy that simulates the 
generalization process by jointly learning the parameters of the encoders and the attention mechanism. As 
shown in Figure 1, our method consists of the following components: 

\begin{itemize}
\item An augmentation mechanism that transforms a sampled graph into one contextual view and two topological views. The augmentations are applied to the initial node feature and graph structure.
\item An encoder consisting of two dedicated GNNs, i.e., graph encoders, and an MLP for the contextual and topological views, respectively, and an attention mechanism to aggregate the learned features.
\item A meta-learning mechanism to jointly learn the parameters of the dedicated encoders and attention model based on error signals from the query set.
\end{itemize}

\subsection{Augmentations}
\label{augmentation}
Recent works on self-supervised learning on graphs suggest that contrasting graph augmentations allows encoders to learn rich node/graph
representations \cite{pmlr-v119-hassani20a}. In this work, we are specifically interested in task-specific and domain-agnostic views of graphs to help the meta-learner to gradually accumulate domain-agnostic 
knowledge while utilizing task-specific information for fast adaptation. We use both feature-space and 
structure-space augmentations as follows.
For the contextual view, we considered three feature-space augmentations on the initial node features 
including: (1) Heterogeneous feature augmentation \cite{duan2012learning} where the initial feature and its 
projection by a linear layer are concatenated and padded to a predefined dimension, (2) Deep set 
\cite{zaheer2017deep} approach in which we considered the initial node feature space as a set, projecting each 
dimension independently to a new space using a linear layer, and aggregating them by a 
permutation-invariant function. This augmentation can capture the shared information across tasks with overlapping features when the alignment among the features is not available. (3) Simple padding of the 
features to a predetermined dimension. Surprisingly, we observed that the simplest augmentation achieves 
better results. We speculate this is because the tasks are not sharing overlapping features.

For the topological view, we apply one feature-space and one structure-space augmentation. In the feature-space augmentation, we replace the task-dependent node features with sinusoidal node degree encodings which allow the model to extrapolate to node degrees greater than the ones encountered 
during meta-training stage \cite{vaswani2017attention}. Because node degrees are universal properties of graph 
nodes, encoding a graph with such initial features will capture task-agnostic geometric structure of the graph. We also 
use graph sub-sampling to keep the degree distribution in a similar order of magnitude across domains.
For the structure-space augmentation, we compute graph diffusion to provide a global view of the graph's 
structure. We used Personalized PageRank (PPR) \cite{page_1999_stanford}, a specific instantiation of the 
generalized graph diffusion. We compute the eigenvalues of the diffusion matrix, sort them in a descending 
order, and select the top-k eigenvalues as the structural representation. We also experimented 
heat kernel diffusion, eigen values of normalized graph Laplacian, and shortest path matrix, and found that diffusion produced better results. 

\subsection{Encoders}
\label{encoder}

Assume a support set $D_i^{sup}=[g_1, g_2,...,g_N]$ of $N$ graphs belonging to a randomly sampled task $i$. 
Augmenting each graph $g$  produces three views: a contextual view represented as graph $g_{c}=(\textbf{A}, \textbf{X}) $ where 
$\textbf{A}\in\lbrace 0, 1 \rbrace^ {n\times n}$ and $\textbf{X}\in \mathbb{R}^{n \times d_{x}}$ denote the adjacency matrix and the task-specific node features, a 
topological view represented as graph $g_{g}=(\textbf{A}, \textbf{U}) $ where $\textbf{U}\in \mathbb{R}^{n \times d_{u}}$ denotes sinusoidal node degree 
encodings, and another topological view represented as vector $\textbf{z}\in \mathbb{R}^{d_{z}}$ denoting the sorted eigenvalues of the 
corresponding diffusion matrix $\textbf{S}\in \mathbb{R}^ {n\times n}$. Our framework allows various choices of  network architecture 
without any constraints. For encoding graph-structured views, we opted for expressive power and adopted 
graph isomorphism network (GIN) \cite{xu_2019_iclr}. The $k^{th}$ layer of our graph encoder consists of a GIN 
layer followed by a feature-wise transformation layer (FWT) \cite{Tseng2020cross} and a swish activation. FWT 
layer simulates various feature distributions under different domains: $h_v^{(k)}=\gamma^{(k)} \times h_v^{(k)} + \beta^{(k)}$ where 
$\gamma \sim \mathcal{N}\left(1, \text{softplus} (\theta_\gamma)\right)$ and $\beta \sim \mathcal{N}\left(0, \text{softplus} (\theta_\beta)\right)$. $\theta_\gamma, \theta_\beta$  are the standard deviations of the Gaussian 
distributions for sampling the affine transformation parameters. 

We use a dedicated graph encoder for each view:  $g_\theta(.): \mathbb{R}^{n \times d_x} \times \mathbb{R}^{n \times n} \longmapsto \mathbb{R}^{n \times d_h}$ and 
$g_\phi(.): \mathbb{R}^{n \times d_u} \times \mathbb{R}^{n \times n} \longmapsto \mathbb{R}^{n \times d_h}$ resulting in two sets of node representations $\mathbf{H}^x$, $\mathbf{H}^u \in \mathbb{R}^{n \times d_h}$
corresponding to the contextual and the topological views of the sampled graph. For each view, we aggregate 
the node representations into a graph representation using a pooling (readout) function$
\mathcal{R}(.) : \mathbb{R}^{n \times d_h} \longmapsto \mathbb{R}^{d_h}$. 
We experimented with global soft attention pooling \cite{li2015gated}, jumping knowledge network 
\cite{Xu_2018_icml}, and summation and mean pooling layers, and found that they produce similar results. Therefore, we opted for simplicity and used a simple mean pooling layer. This results in two graph 
representations: $\textbf{h}^x, \textbf{h}^u \in \mathbb{R}^{d_h}$. We also feed the topological view from the eigenvalues of the graph diffusion 
into a projection head $f_\psi(.): \mathbb{R}^{d_z}\longmapsto \mathbb{R}^{d_h}$, modeled as an MLP resulting in the third representation: $\textbf{h}^z \in \mathbb{R}^{d_h}$.

To aggregate the learned representations, we feed the concatenation of the learned representations into an 
attention module $f_\omega(.): \mathbb{R}^{3 \times d_h}\longmapsto \mathbb{R}^{3}$ that generates attention scores for each representation. The attention 
module is modeled as a single-layer MLP followed by a softmax function:

\begin{equation}
\label{eq:att}
\alpha= \text{Softmax}\bigg(\text{ReLU}\bigg(\bigg[ \textbf{h}^x \parallel \textbf{h}^u \parallel \textbf{h}^z \bigg]\mathbf{W_1}\bigg)\mathbf{W_2}\bigg) 
\end{equation}

where $\mathbf{W_1}\in \mathbb{R}^{(3 \times h_d) \times h_d}$ and $\mathbf{W_2}\in \mathbb{R}^{h_d \times 3}$ are network parameters. The attention scores are then used to 
aggregate the learned features into a final graph representation.

The attention mechanism gates the representations and decides if the model should rely more on contextual or 
topological representations. If the samples are from a task that is similar to seen tasks, the model will pay more 
attention to contextual representation whereas if there is a drastic shift in feature space, the model will rely 
more on geometric representations. We assume that the target domain is not available during training and 
hence there is no information in advance about whether there are shared features among tasks. If there is, the 
attention module will pass the shared contextual information through, otherwise it will not attend to the 
contextual features and will let the learner to learn them from scratch during the meta-test adaptation phase. 
Hence, rather than naively throwing the information away and learning from scratch, we let the model decide if it 
can use the information. It is noteworthy that we are not introducing a new meta-learning framework. Instead we are 
introducing an encoder with attention module that can seamlessly be integrated into any meta-learning 
framework. As an example, we show the training procedure of the encoder within a mini-batch of tasks using 
prototypical approach \cite{snell2017prototypical} in Algorithm \ref{algo}. Depending on the meta-learner, the 
aggregated representation can be then fed into a linear classifier or a non-parametric classifier such as a 
prototypical classifier. 

\begin{algorithm}[t]
\caption{Training the proposed encoder with prototypical approach for one mini-batch of tasks. $\textbf{A}_g, 
\textbf{S}_g, \textbf{X}_g, \textbf{U}_g$ denote adjacency, diffusion, node features, and node degree encodings of graph $g$.}
\label{algo}
\SetAlgoLined \DontPrintSemicolon
\KwIn{Concatenation operator $\parallel$, readout function $\mathcal{R}$, eigenvalue function $\mathcal{E}$, 
prototype function $\mathcal{P}$, prototypical loss $\mathcal{L}$, encoders $g_\theta$, $g_\phi$, $f_\psi$, 
attention module $f_\omega$, and meta-training task batch $\{\mathcal{T}_j|\mathcal{T}_j= D^{s}_{\mathcal{T}_j} 
\cup D^{q}_{\mathcal{T}_j} \}_{j=1}^{N}$ }

\For{$\mathcal{T}$ in a task batch $\{\mathcal{T}_j \}_{j=1}^{N}$}
{
$[\textbf{H}^s, \textbf{H}^q, \mathcal{Y}^s, \mathcal{Y}^q] \gets \emptyset$ \\
\For{$(g, y)$ in $D^{s}_{\mathcal{T}} \cup D^{q}_{\mathcal{T}}$}
{
$\textbf{h}_x \gets \mathcal{R}\left(g_\theta\left(\textbf{A}_g, \textbf{X}_g \right) \right)$ \\
$\textbf{h}_u \gets \mathcal{R}\left(g_\phi\left(\textbf{A}_g, \textbf{U}_g \right) \right)$ \\
$\textbf{h}_z \gets  f_\psi\left( \mathcal{E}\left(\textbf{S}_g \right) \right)$ \\
$\alpha \gets f_\omega \left( \left[ \textbf{h}_x \parallel \textbf{h}_u \parallel \textbf{h}_z \right] \right)$ \\
$\textbf{h} \gets \alpha_0 \textbf{h}_x + \alpha_1 \textbf{h}_u + \alpha_2 \textbf{h}_z $\\
\eIf{$g \in D^{s}_{\mathcal{T}}$}{
$\textbf{H}^s \gets \textbf{H}^s \parallel\textbf{h}$ , $\mathcal{Y}^s \gets \mathcal{Y}^s \parallel y$
}
{
$\textbf{H}^q \gets \textbf{H}^q \parallel\textbf{h}$ , $\mathcal{Y}^q \gets \mathcal{Y}^q \parallel y$
}
}
$\textbf{C}_\mathcal{T} \gets \mathcal{P}\left( \mathbf{H}^s, \mathcal{Y}^s \right)$
$\textbf{L}_\mathcal{T} \gets \textbf{L}_\mathcal{T} + \mathcal{L}\left(\textbf{C}_\mathcal{T}, \mathbf{H}^q, \mathcal{Y}^q \right) $
}$\left[\theta, \omega, \phi, \psi \right] \gets \left[\theta, \omega, \phi, \psi \right] - \gamma\nabla_{\left[\theta, \omega, \phi, \psi \right]} \frac{1}{N \times |D^q|} \sum\limits_{j=1}^N{ \left[ \textbf{L}_{\mathcal{T}_j}\right]}$
\end{algorithm}

\begin{table*}
\setlength{\tabcolsep}{2.4pt}
\caption{Statistics of the proposed benchmarks for heterogeneous cross-domain few-shot graph classification.} \label{table:stat}
\begin{center}
\begin{small}
\begin{sc}
\begin{tabular}{cc|ccc|ccc|ccc}
\toprule
\multicolumn{2}{c}{\textbf{Meta-Domain}} & \multicolumn{3}{c}{\textbf{$|$Task$|$}} & \multicolumn{3}{c}{\textbf{Avg. on target}} & \textbf{$|$shot$|$} & \textbf{$|$class$|$} & \textbf{$|$query$|$} \\
\cmidrule{1-8}
Source & Target & Train & Dev & Test & Node & Edge & Feature & \\
\midrule
Molecules & Molecules            & 169 & 5 & 18 & 26.6 $\pm$ 15.7 & 28.6$\pm$16.6 & 18.1 $\pm$  18.7 & 1,5,10,20 & 2 & 50 \\
Molecules & Bioinformatics       & 187 & 5 & 24 & 79.2 $\pm$ 58.5 & 406.6$\pm$300.3  & 19.8 $\pm$  15.1 & 1,5,10,20 & 2 & 50  \\
Molecules & Social Networks      & 187 & 5 & 12 & 54.1 $\pm$ 58.8 & 98.1$\pm$117.9 & 0 & 1,5,10,20 & 2 & 50 \\
\bottomrule
\end{tabular}
\end{sc}
\end{small}
\end{center}
\end{table*}

\begin{table*}
\caption{Mean and Standard Deviation of meta-test accuracy on Bioinformatics benchmark after ten runs.}
\label{table:graph}
\begin{center}
\begin{footnotesize}
\begin{sc}
\begin{tabular}{clccccc}
\toprule
\multicolumn{2}{c}{\textbf{method}} & \textbf{1-shot} & \textbf{5-shot} & \textbf{10-shot} &\textbf{ 20-shot} \\
\midrule
& Empirical Upper Bound  & \multicolumn{4}{c}{\textbf{66.78 $\pm$ 10.30}} \\
\midrule
\multirow{3}{*}{\begin{turn}{90}sup\end{turn}} 
& GCN  & 54.88 $\pm$ 7.55   & 55.05 $\pm$ 8.85  & 55.03 $\pm$ 8.91  & 54.99 $\pm$ 8.82  \\
& GAT  & 54.75 $\pm$ 8.85   & 54.69 $\pm$ 8.90  & 54.76 $\pm$ 8.97 & 54.63 $\pm$ 8.94 \\
& GIN  & 55.37 $\pm$ 9.83   & 55.52 $\pm$ 9.79  & 55.47 $\pm$ 9.89   & 55.52 $\pm$ 9.65 \\
\midrule
\multirow{4}{*}{\begin{turn}{90}uns\end{turn}} 			
& InfoGraph  &  54.00 $\pm$ 6.65 & 53.67 $\pm$ 7.35 & 54.42 $\pm$ 6.41 & 54.96 $\pm$ 7.63  \\ 
& MVGRL      & 57.12 $\pm$ 7.75 & 57.25 $\pm$ 9.04 & 57.17 $\pm$ 8.01 & 57.54 $\pm$ 8.06\\
& GSFE       & 52.84 $\pm$ 6.71 & 52.96 $\pm$ 7.82 & 53.06 $\pm$ 7.64 & 53.16 $\pm$ 7.87   \\ 
& GCC        & 53.11 $\pm$ 6.51 & 53.17 $\pm$ 6.43 & 53.18 $\pm$ 7.18 & 53.35 $\pm$ 7.64   \\			
\midrule
\multirow{11}{*}{\begin{turn}{90}meta\end{turn}}
& MatchNet  &  54.83  $\pm$  7.66 &  55.62  $\pm$  7.60 & 55.92  $\pm$  6.67 & 56.04  $\pm$  7.78 \\
& ProtoNet  & 54.71  $\pm$  8.86  & 55.75  $\pm$  7.84  & 55.96  $\pm$  6.73 & 55.50  $\pm$  9.65 \\
& RelationNet  & 54.93 $\pm$   8.55  & 55.92 $\pm$   8.69  & 56.02 $\pm$   7.69 & 56.15 $\pm$   7.81 \\
& MAML  & 53.83 $\pm$  9.62 & 54.46  $\pm$  6.77  & 54.50  $\pm$  8.77  & 54.79  $\pm$  8.90 \\
& MetaSGD  & 53.83  $\pm$  8.79 &  54.21  $\pm$  7.70 & 54.67  $\pm$  9.72 & 54.71  $\pm$  7.90 \\
& MetaSpecGraph & $-$  & 55.47  $\pm$  7.79  & 55.82  $\pm$  8.91	 & 55.97  $\pm$  8.89 \\
\cmidrule{2-6}
& MatchNet + Our Encoder         & \textbf{59.14 $\pm$ 7.00}  & \textbf{59.19 $\pm$ 9.77}   & 59.22 $\pm$ 8.72  & 59.56 $\pm$ 6.97  \\			
& ProtoNet + Our Encoder           & 57.17 $\pm$ 7.76   & 57.58 $\pm$ 8.89   & 57.79 $\pm$ 8.76  & 58.17 $\pm$ 7.88  \\
& RelationNet + Our Encoder      & 58.83 $\pm$ 8.03  & 58.83 $\pm$ 9.68  & \textbf{59.29 $\pm$ 7.87}  & \textbf{59.82 $\pm$ 7.93}  \\
& MAML + Our Encoder               & 56.00 $\pm$ 8.74   & 56.21 $\pm$ 7.76  & 56.37 $\pm$ 8.81   & 57.04 $\pm$ 7.85 \\
& MetaSGD + Our Encoder           & 55.08 $\pm$ 8.67      & 56.12 $\pm$  8.23      & 57.08 $\pm$   8.86      &  57.33$\pm$8.86\\
\bottomrule
\end{tabular}
\end{sc}
\end{footnotesize}
\end{center}
\end{table*}

\begin{table*}
\caption{Mean and Standard Deviation of meta-test accuracy on Social Networks benchmark after ten runs.}
\label{table:graph}
\begin{center}
\begin{footnotesize}
\begin{sc}
\begin{tabular}{clccccc}
\toprule
\multicolumn{2}{c}{\textbf{method}} & \textbf{1-shot} & \textbf{5-shot} & \textbf{10-shot} &\textbf{ 20-shot} \\
\midrule
& Empirical Upper Bound  & \multicolumn{4}{c}{\textbf{72.35 $\pm$ 12.38}} \\
\midrule
\multirow{3}{*}{\begin{turn}{90}sup\end{turn}} 
& GCN  & 60.51 $\pm$ 10.54 & 60.54 $\pm$ 10.18  & 60.33 $\pm$ 10.19 & 60.35 $\pm$ 10.52  \\
& GAT  & 61.32 $\pm$ 10.31 & 61.37 $\pm$ 10.20  & 61.17 $\pm$ 10.17 & 61.51 $\pm$ 10.13 \\
& GIN  &  62.11 $\pm$ 10.11 & 62.98 $\pm$ 10.04 &  63.27 $\pm$ 10.05 & 63.24 $\pm$ 9.28 \\
\midrule
\multirow{4}{*}{\begin{turn}{90}uns\end{turn}}
& InfoGraph  & 61.92 $\pm$ 9.84  & 62.25 $\pm$ 7.12 & 62.58 $\pm$ 8.57  &  62.58 $\pm$ 7.32   \\ 
& MVGRL      & 63.00 $\pm$ 10.70 & 63.75 $\pm$ 11.17 & 63.25 $\pm$ 11.69 &  63.75 $\pm$ 11.99  \\
& GSFE  & 60.38 $\pm$ 9.74  & 60.45 $\pm$ 9.62  & 60.46 $\pm$ 9.95  &  60.55 $\pm$ 9.11  \\ 			
& GCC      & 60.61 $\pm$ 9.55 & 60.73 $\pm$  9.74 &  60.81 $\pm$ 9.61 &  60.98 $\pm$ 9.97  \\			
\midrule
\multirow{11}{*}{\begin{turn}{90}meta\end{turn}}
& MatchNet  & 62.25  $\pm$  9.14  & 62.42  $\pm$  9.98  & 62.92  $\pm$  9.22 &  63.33  $\pm$  9.92 \\
& ProtoNet  & 60.50  $\pm$  9.05  & 61.25  $\pm$  10.01  & 61.75  $\pm$  9.06 & 63.50  $\pm$  9.97 \\
& RelationNet  & 61.25 $\pm$   10.04  & 61.08 $\pm$   10.25 & 61.83 $\pm$   10.17 & 62.00 $\pm$   10.17 \\
& MAML  & 58.33  $\pm$  9.05  & 58.75  $\pm$  10.84  & 59.00  $\pm$  9.87 & 60.17  $\pm$  10.99 \\
& MetaSGD  & 59.25  $\pm$  10.15  & 59.33  $\pm$  9.50  & 59.83  $\pm$  9.89 & 60.25  $\pm$  9.88 \\
& MetaSpecGraph  & $-$  & 62.55  $\pm$  9.79  & 62.74  $\pm$  9.91 & 63.73  $\pm$  9.89 \\
\cmidrule{2-6}
& MatchNet + Our Encoder & \textbf{67.40 $\pm$ 10.37} & \textbf{67.71 $\pm$ 9.24} &  \textbf{68.34 $\pm$ 9.09} &  68.90 $\pm$ 9.18  \\
& ProtoNet + Our Encoder  & 66.42 $\pm$ 10.21  & 66.92 $\pm$ 10.55  & 67.00 $\pm$ 9.13 & 67.50 $\pm$ 10.46 \\
& RelationNet + Our Encoder                            & 66.75 $\pm$ 9.89 & 67.08 $\pm$ 10.80 & 67.33 $\pm$ 10.33 &   \textbf{69.67 $\pm$ 10.05}  \\
& MAML + Our Encoder                  & 62.25 $\pm$ 10.12 & 64.75 $\pm$ 10.79 &  66.58 $\pm$ 10.93 & 66.83 $\pm$ 10.21 \\
& MetaSGD + Our Encoder                     & 61.92$\pm$ 9.56  & 63.17 $\pm$ 10.51  & 64.42 $\pm$  10.16  & 64.83 $\pm$  10.06  \\
\bottomrule
\end{tabular}
\end{sc}
\end{footnotesize}
\end{center}
\end{table*}

\section{Benchmarks}
We also introduce three new few-shot graph classification benchmarks with fixed meta-- train/val/test 
splits constructed from publicly available graph datasets. In all benchmarks, the source meta-domain consists of molecule classification tasks. We made this 
decision because most of the available graph classification datasets are molecule datasets and hence using 
them as the source meta-domain can provide sufficient tasks during meta-training. The target meta-domains 
are molecule, bioinformatics, and social networks. Note that although both source and target meta-domains in 
the first benchmark pertain to molecule, the tasks differ in both feature and class spaces. The process of 
creating these benchmark was as follows.

We collected all the datasets from TUDataset \cite{morris2020tudata} and OGB \cite{hu2020open}. We kept graphs 
with maximum node degree of 50, maximum node feature dimension of 100, and graphs with a minimum of 
two nodes and one edge and ignored the rest. We also filtered out graphs with disconnected components. For 
graphs with more than 500 nodes, we sorted the nodes by their harmonic centrality \cite{boldi2014axioms} in a 
descending order and used the sub-graphs containing the top 500 nodes. For multi-task datatsets, we drew 
samples from each task without replacement and split them into several single-task datasets without sharing 
any data samples. Because the majority of datasets in TUDataset and OGB are binary 
classification tasks, we opted for a ``k-shot 2-way'' setting and split the few remaining multi-class datasets into 
binary datasets by sampling without replacement. We then randomly selected 20 and 50 samples per class as 
support and query sets, respectively. 

For the second and third benchmarks, we used the processed datasets from bioinformatics and social network 
categories as the meta-testing tasks. For the first benchmark, we split the tasks such that if a task is originated 
from a multi-task or a multi-class dataset, it is allocated to a split with all of the tasks that also originated from 
the same dataset. The statistics of the proposed benchmarks are shown in Table \ref{table:stat}. We believe these 
benchmarks will help the community to drive further advances in heterogeneous meta-learning. 

\section{Experimental Results}
We exhaustively perform empirical evaluations to answer the following questions: (1) what is the the empirical 
upper bound of the classification accuracy on the meta-test set for the benchmarks? (2) does any knowledge 
transfer occur across meta-domains? If not, does negative transfer occur? (3) How well does pre-training based 
on contrastive methods perform? (4) How do metric-based meta-learning methods perform compared to 
optimization-based methods? (5) what is the effect of using the proposed encoder?

To estimate the empirical upper bound of the classification accuracy, we approximated how fully-supervised 
models with access to all data would perform. To accomplish this, we used all the data available within the 
datasets from which the few-shot tasks in the meta-testing phase are sampled and trained one classifier per 
meta-testing task in a fully supervised fashion. We evaluated the classifiers on the same query sets used in the 
few-shot setting, i.e., query samples are fixed, but the support set is replaced with all data available in the 
original dataset. To investigate knowledge transfer, we trained independent classifiers on the support set of the 
meta-testing tasks and evaluated them on their corresponding query set. Negative transfer occurs when the 
performance on meta-testing tasks is negatively affected by knowledge transferred from meta-training tasks 
\cite{pan2009survey}. Therefore, if the independently trained models outperform models that utilize knowledge 
transfer, it implies negative transfer. We also investigated two strategies to utilize knowledge transfer: 
pre-training using contrastive methods and learning-to-learn strategies. For contrastive methods, we 
aggregated support and query samples from the meta-training tasks into a unified dataset, pre-trained a GNN, 
and then fine-tuned it to the meta-test tasks. Finally, we investigated the effectiveness of the proposed encoder 
by comparing the performance of the meta-learning models with and without using it. 

We report the mean classification accuracy with standard deviation over query samples of the meta-testing 
tasks after ten runs. For estimating empirical upper bound accuracy, we used a GIN \cite{xu_2019_iclr} classifier. 
For independent classifiers, we used GCN \cite{kipf_2017_iclr}, GAT \cite{velickovic_2018_iclr}, and GIN layers. 
For contrastive methods, we used GCC \cite{GCC2020}, GSFE \cite{hu2019pre}, InfoGraph 
\cite{Sun2020InfoGraph}, and MVGRL \cite{pmlr-v119-hassani20a}. During the meta-testing phase, we first 
fine-tuned the encoder on the support set using the same contrastive approach, and then trained an MLP on 
the graph representations. We found that concatenating the graph representation produced by the contrastive 
method with a mean pooling of initial node features improves the performance. For meta-learning methods, 
we used three metric-based techniques including matching network \cite{vinyals2016matching}, prototypical 
network \cite{snell2017prototypical}, and relation network \cite{sung2018learning}, and two optimization-based 
methods including MAML \cite{finn2017model} and MetaSGD \cite{li2017meta}. For optimization-based methods, 
we used a linear classifier in both the meta-training and meta-testing phases whereas for metric-based 
methods, we found that using a cosine distance classifier \cite{chen2018a} for adaptation achieves better  
performance. We trained the meta-learning models with and without our proposed encoder. We also trained a 
specialized meta-learning approach for graph classification that uses spectral information 
\cite{Chauhan2020Few}. The results are shown in Tables 2 and 3 (for Molecule benchmark results, 
hyper-parameters and other details refer to Appendix).

Results suggest that: (1) On all three benchmarks, there exists underlying knowledge that can be transferred. 
This is validated by the observation that both meta-learning and contrastive approaches outperform naive 
classifiers. (2) Contrastive approaches are competitive with meta-learning methods without the proposed 
encoder. As an example, on 20-shot bioinformatics benchmark, MVGRL outperforms the best performing 
meta-learning method by 1.57\% absolute accuracy. (3) Coupling metric-based meta-learning methods with 
our proposed encoder significantly enhances performance. As an instance, on 1-shot setting, the best 
meta-learning methods coupled with our encoder outperform the best results achieved by regular 
meta-learning methods by 3.28\%, 4.29\%, and 5.17\% absolute accuracy on molecules, bioinformatics, and 
social network benchmarks, respectively. (4) RelationNet coupled with our encoder and only trained with 20 
examples is only 4.46\%, 6.96\%, and 2.68\% less accurate than fully-supervised models trained on all 
available data of molecules, bioinformatics, and social network benchmarks, respectively. Note that some of 
these datasets have tens of thousands training samples. (5) We get the most improvement when we transfer 
knowledge from molecular meta-training to social network meta-testing. This is because social network tasks 
do not contain any initial node features, and hence classifying them completely depends on task-agnostic 
geometric features. This suggests that our encoder is able to learn expressive geometric representations on one 
domain and generalize to another domain. 

\subsection{Ablation Study}
\subsubsection{Effect of Views.}
To investigate the effect of views and the proposed encoder, we run the experiments with: (1) Only node features \textbf{X} which is equivalent to traditional meta-learning, (2) a combination of node features \textbf{X} and node degree encodings \textbf{U}, (3) a combination of node features \textbf{X} and eigenvalues of diffusion matrix \textbf{Z}, and (4) combination of all the features. The result (see Tables 6-8 in Appendix) suggest that: (1) both topological views contribute complementary enhancements to the performance, and (2) the proposed encoder has a better performance when coupled with metric-based methods compared to optimization-based methods. Also we observe that  the diffusion based view has a slightly larger effect on performance. We speculate this is because of the inductive bias that diffusion provides (it directly encodes graph encoding from global information).

\subsubsection{Why topological properties are task-agnostic?}
Node degree distributions are different across domains (e.g., social networks and molecules). We are addressing
this by a few tricks: (1) we use sinusoidal node degree encodings which allow the model to extrapolate to unseen node
degrees, (2) we use graph sub-sampling to keep the degree distribution in a similar order of magnitude, and most 
importantly, (3) we are interleaving FWT and GNN layers to address the distribution shift. This allows us to first 
relax the heterogeneity by relying on
topological signals and then use the mentioned tricks to control the
distribution shifts that may occur. It is noteworthy that we found the encodings from diffusion to be less sensitive to the shifts in graph size and node degrees distributions. 

\subsubsection{Effect of Meta-Learning Approach.}
We considered optimization-based and metric-based approaches. Results suggest that without using our encoder, metric-based approaches are competitive with contrastive methods and also show some degree of 
knowledge transfer, whereas optimization-based approaches are outperformed by independent classifiers.
That suggests that optimization-based approaches lead to negative knowledge transfer, confirming the observation in \cite{guo2020broader}. When coupled with our encoder, optimization-based approaches become 
competitive with contrastive methods and show some knowledge transfer but are still outperformed by
metric-based approaches (see Appendix). As suggested in \cite{huang2020graph}, this is likely because metric-based approaches leverage the inductive bias between representation and labels to effectively propagate scarce 
label information.

\subsubsection{Effect of Task Encoding.}
We attempted to explicitly condition the attention module on the task encodings as follows. We pretrained an 
encoder using MVGRL and encode the support set of each task. We then aggregated the representation into a 
task encoding and stored meta-training task encodings within a memory-bank. We trained the attention 
module by feeding it with the graph views, the task encodings, and the encodings of the top-3 similar tasks. 
During meta-testing, we retrieve the top-3 similar tasks from the memory bank. Surprisingly, we found that 
using task encodings produces nearly identical results to not conditioning the attention module on task 
encodings.

\section{Conclusion}
We investigated the problem of cross-domain few-shot graph classification by introducing three 
new benchmarks. We performed exhaustive experiments using independent classifiers, contrastive methods, 
and meta-learning strategies. We also introduced a simple yet powerful multi-view graph encoder with an 
attention-based aggregation mechanism for better knowledge transfer and adaptation. We showed that 
metric-based meta-learning approaches coupled with the proposed encoder achieve the best performance 
across all benchmarks. We also showed that optimization-based meta learning methods struggle in 
cross-domain setting. In future work, we plan to investigate black-box and hybrid adaptations 
of meta-learning strategies.

\bibliography{aaai22}

\newpage
\section{Appendix}
\subsection{Datasets, Pre-processing, \& Implementation}
The statistics of the used datasets is reported in Tables 1-3. Mutli-task column indicates whether the original dataset is a multi-task dataset whereas task column 
denotes the number of tasks extracted from the dataset for few-shot setting. Class column indicates the number of classification categories in the original dataset.
As shown, most of the datasets are for binary classification problems and also molecule datasets are the most frequent ones. This is why we choose to set the
molecules meta-domain as the meta-training domain and set the problem to ``k-way 2-shot''.

We addressed the datasets without initial node features as follows. If a dataset does not carry initial node features but contains node labels, we use the one-hot 
encoding of the node labels as the initial features. Otherwise, we initialize the node features with a 16-dimensional vector of ones. We also normalize the initial
node features row-wise.

We implemented the experiments using PyTorch, and used Pytorch Geometric and Learn2Learn libraries to implement graph encoders and optimization-based
meta-learning algorithms. The experiments are run on four RTX 6000 GPUs where on average one epoch of training optimization-based algorithms takes 45 
seconds and training one epoch of metric-based methods takes about 6 seconds. 

\subsection{Hyper-Parameters}
We initialize the parameters using Xavier initialization and train all the model using Adam optimizer. For all the experiments, we fix the number of epochs to 1000, 
the size of hidden dimension to 256, size of sinusoidal node degree embeddings to 32, and the number of eigenvalues of the diffusion matrix to 128. During the 
meta-training, we use a cosine scheduler to schedule the learning rate, and also use early stopping with patience of 30. We also use a meta-batch of 16 tasks. We
choose the number of task steps, number of adaptation steps, task learning rate, and adaptation learning rate from [10, 25, 50], [10, 25, 50],   [0.001, 0.01], and
[0.01, 0.1], respectively. For the encoders we choose the number of GNN and MLP layers from [1, 2, 3]. To prevent the models from over-fitting, we use dropout
with probability of 0.6, and feature-wise transform after the linear layers. We found that sampling the hyper-parameters of the feature-wise transform layers 
produces good results in our heterogeneous setting. We set the teleport probability for computing the diffusion to 0.2. For MAML and MetaSGD, we also experimented 
with first and second order optimization settings. Finally, for matching network, we used a bidirectional LSTM to generate fully conditional embeddings of the support 
set and an attentional LSTM to generate fully conditional embeddings of the query set. The selected hyper-parameters for 20-shot setting are shown in Table 4.

\subsection{Effect of Views \& Encoder}
To investigate the effect of views and the proposed encoder, we run the experiments with: (1) Only initial node features \textbf{X} which is equivalent to traditional 
meta-learning, (2) initial node features \textbf{X} and node degree features \textbf{U}, (3) initial node features \textbf{X} and eigenvalues of diffusion matrix \textbf{Z}, and (4) all the mentioned
features. Setting 2-4 are essentially adding the proposed encoder to the meta-learning algorithms. The result shown in Tables 6-8 suggest that: (1) both geometric views
contribute complementary enhancements to the performance, and (2) the proposed encoder has a better performance when coupled with metric-based methods 
compared to optimization-based methods.

\section{subsection}
Diffusion is formulated as Eq. (\ref{eq:ggd}) where $\textbf{T} \in \mathbb{R}^{n\times n}$ is the generalized transition matrix and $\Theta$ is the weighting coefficient which determines the ratio of 
global-local information. Imposing $\sum_{k=0}^{\infty} \theta_k=1$, $\theta_k \in [0,1]$, and $\lambda_i \in [0,1]$ where $\lambda_i$ are eigenvalues of \textbf{T}, guarantees convergence. Diffusion is computed once 
using fast approximation and sparsification methods.

\begin{equation}
    \label{eq:ggd}
	\textbf{S}=\sum_{k=0}^{\infty}{\Theta_k \textbf{T}^k} \in \mathbb{R}^{n\times n}
\end{equation}

Given an adjacency matrix $\textbf{A}\in \mathbb{R}^{n\times n}$ and a diagonal degree matrix $\textbf{D}\in \mathbb{R}^{n\times n}$, Personalized PageRank (PPR) and heat kernel , i.e., two instantiations of the 
generalized graph diffusion, are defined by setting $\mathbf{T}=\textbf{AD}^{-1}$, and $\theta_k=\alpha(1-\alpha)^k$ and $\theta_k= e^{-t}t^{k}/{k!}$, respectively, where $\alpha$ denotes teleport probability in a 
random walk and $t$ is diffusion time. 

\begin{table*} \vskip -0.15in
\setlength{\tabcolsep}{3pt}
\caption{Statistics of the datasets used in molecules meta-domain.}
\label{table:graph}
\begin{center}
\begin{footnotesize}
\begin{sc}
\begin{tabular}{lcccccccc}
\toprule
\textbf{Dataset} & \textbf{Source} & \textbf{Multi-Task} & \textbf{$|$class$|$} & \textbf{$|$graph$|$} &\textbf{Avg. Node} &\textbf{Avg. Edge} &  \textbf{$|$task$|$} & \textbf{split}\\
\midrule
AIDS   & TU & N & 2 &  2000 & 15.69 & 16.20 & 1 & meta-train\\
BZR	 & TU & N &  2 &  405	& 35.75 & 38.36	& 1 & meta-train\\
COX2	 & TU & N &  2 &  467 & 41.22 & 43.45& 1 & meta-train\\
DHFR	 & TU & N &  2 &  467 & 42.43 & 44.54 & 1 & meta-train\\	
MCF-7	 & TU & N &  2 &  27770 &  26.39 &  28.52 & 1 & meta-train\\
MCF-7H	& TU & N &  2 &  27770 &  47.30 &  49.43 & 1 & meta-train\\
MOLT-4	 & TU & N &  2 &  39765 &  26.09 &  28.13 & 1 & meta-train\\
MOLT-4H	& TU & N &  2 &  39765 &  46.70 &  48.73 & 1 & meta-train\\
MUTAGENICITY & TU & N &  2 &  4337 &  30.32 &  30.77 & 1 & meta-train\\
NCI1	& TU & N &  2 &  4110 &  9.87 &  32.30 & 1 & meta-test \\
NCI109 & TU & N &  2 &  4127 &  29.68 &  32.13 & 1 & meta-test \\
P388	& TU & N &  2 &  	41472 &  22.11 &  23.55 & 1 & meta-train\\
P388H	& TU & N &  2 &  	41472 &  40.44 &  41.88 & 1 & meta-train\\
PC-3	& TU & N &  2 &  	27509 &  26.35 &  28.49 & 1 & meta-train\\
PC-3H	& TU & N &  2 &  	27509 &  47.19 &  49.32 & 1 & meta-train\\
PTC-FM	& TU & N &  2 &  349 &  14.11 &  14.48 & 1 & meta-test\\
PTC-FR	& TU & N &  2 &  	351 &  14.56 &  15.00 & 1 & meta-test\\
PTC-MM	& TU & N &  2 &  	336 &  13.97 &  14.32 & 1 & meta-test\\
PTC-MR	& TU & N &  2 &  	344 &  14.29 &  14.69 & 1 & meta-test\\
SF-295	& TU & N &  2 &  40271 &  26.06 &  28.08 & 1 & meta-train\\
SF-295H	& TU & N &  2 &  40271 &  46.65 &  48.68 & 1 & meta-train\\
SN12C	& TU & N &  2 &  40004 &  26.08 &  28.11 & 1 & meta-train\\
SN12CH	& TU & N &  2 &  40004 &  46.69 &  	48.71 & 1 & meta-train\\
SW-620	& TU & N &  2 &  	40532 &  26.05 &  28.08 & 1 & meta-val\\
SW-620H	& TU & N &  2 &  	40532 &  46.62 &  48.65 & 1& meta-val\\
Tox21-AhR	 & TU & N &  2 &  	8169 &  18.09 &  18.50 & 1 & meta-test\\
Tox21-AR & TU & N &  2 &  	9362 &  18.39 &  18.84 & 1& meta-test\\
Tox21-AR-LBD	& TU & N &  2 &  	8599 &  17.77 &  18.16 & 1 & meta-test\\
Tox21-ARE	& TU & N &  2 &  7167 &  16.28 &  16.52 & 1 & meta-test\\
Tox21-aromatase	& TU & N &  2 &  	7226 &17.50 &17.79 & 1 & meta-test\\
Tox21-ATAD5	& TU & N &  2 &  	9091 &17.89	& 18.30 & 1 & meta-test\\
Tox21-ER	& TU & N &  2 &  	7697 &17.58& 	17.94 & 1 & meta-test\\
Tox21-ER-LBD	& TU & N &  2 &  	8753 & 18.06 &18.47 & 1 & meta-test\\
Tox21-HSE	& TU & N &  2 &  	8150 &16.72 &17.04 & 1 & meta-test\\
Tox21-MMP	& TU & N &  2 &  	7320 &17.49 &17.83 & 1 & meta-test\\
Tox21-p53	& TU & N &  2 &  	8634 &17.79 &18.19 & 1 & meta-test\\
Tox21-PPAR-GAMMA	& TU & N &  2 &  	8184 &17.23	 &17.55 & 1 & meta-test\\
UACC257	& TU & N &  2 &  	39988 &26.09 &28.12 & 1 & meta-train\\
UACC257H	& TU & N &  2 &  	39988 &46.68 &48.71 & 1 & meta-train\\
YEAST	& TU & N &  2 &  	79601 & 21.54 &22.84 & 1 & meta-val\\
YEASTH	& TU & N &  2 & 79601 & 39.44 & 40.74 & 1 & meta-val\\
MOLHIV & OGB & N &  2 & 41127 & 25.51 & 25.51 & 1 & meta-train\\
MOLBACE & OGB & N &  2 & 1513 & 34.09 & 73.72 & 1 & meta-train\\
MOLBBBP & OGB & N &  2 & 2039 & 24.06 & 51.91 & 1 & meta-train\\
MOLPCBA & OGB & Y &  2 & 437929 & 25.97 & 56.22 & 121 & meta-train\\
MOLCLINTOX & OGB & Y &  2 & 1477 & 26.16 & 55.77 & 1 & meta-val\\
MOLSIDER & OGB & Y &  2 & 1427 & 33.64 & 70.72 & 8 & meta-train\\
MOLTOXCAST & OGB & Y &  2 & 8576 & 18.78 & 38.52 & 17 & meta-train\\
\bottomrule
\end{tabular}
\end{sc}
\end{footnotesize}
\end{center}
\end{table*}

\begin{table*}
\setlength{\tabcolsep}{3pt}
\caption{Statistics of the datasets used in bioinformatics meta-domain.}
\label{table:graph}
\begin{center}
\begin{footnotesize}
\begin{sc}
\begin{tabular}{lcccccccc}
\toprule
\textbf{Dataset} & \textbf{Source} & \textbf{Multi-Task} & \textbf{$|$class$|$} & \textbf{$|$graph$|$} &\textbf{Avg. Node} &\textbf{Avg. Edge} &   \textbf{$|$task$|$} & \textbf{Split} \\
\midrule
DD & TU & N &  2 &1178 &284.32 &715.66  & 1 & meta-test\\
ENZYMES & TU & N &  6 &600 &32.63 & 62.14 & 3 & meta-test\\
PROTEINS & TU & N &  2 &1113& 39.06 & 	72.82 & 1 & meta-test\\
PROTEINS-full & TU & N &  2 &1113& 39.06& 	72.82 & 1 & meta-test\\
PPA	& OGB & N &  37 &  	158100 & 243.42 &4532.19 & 18 & meta-test\\
\bottomrule
\end{tabular}
\end{sc}
\end{footnotesize}
\end{center}
\end{table*}

\begin{table*}
\setlength{\tabcolsep}{3pt}
\caption{Statistics of the datasets used in social networks meta-domain.}
\label{table:graph}
\begin{center}
\begin{footnotesize}
\begin{sc}
\begin{tabular}{lcccccccc}
\toprule
\textbf{Dataset} & \textbf{Source} & \textbf{Multi-Task} & \textbf{$|$class$|$} & \textbf{$|$graph$|$} &\textbf{Avg. Node} &\textbf{Avg. Edge} &   \textbf{$|$task$|$} & \textbf{Split} \\
\midrule
COLLAB & TU & N &  3 & 5000 & 74.49 & 2457.78 &  1 & meta-test\\
DEEZER-EGO-NETS	& TU & N &  2 & 9629&  23.49&  65.25 &  1 & meta-test\\
GITHUB-STARGAZERS	& TU & N &  2 & 12725 &  113.79	&  234.64 &  1 & meta-test\\
IMDB-BINARY	& TU & N &  2 & 1000 & 19.77 & 96.53 &  1 & meta-test\\
IMDB-MULTI	& TU & N &  3 & 1500 & 13.00 & 65.94&  1 & meta-test\\
REDDIT-BINARY	& TU & N &  2 & 	2000 & 429.63 & 497.75&  1 & meta-test\\
REDDIT-MULTI-5K	& TU & N &  5 & 4999 & 508.52 & 594.87&  1 & meta-test\\
REDDIT-MULTI-12K& TU & N &  11 & 11929 & 391.41 & 456.89&  3 & meta-test\\
REDDIT-THREADS& TU & N &  2 & 203088 & 23.93 & 24.99&  1 & meta-test\\
TWITCH-EGOS	& TU & N &  2 & 127094 & 29.67 & 86.59&  1 & meta-test\\
\bottomrule
\end{tabular}
\end{sc}
\end{footnotesize}
\end{center}
\end{table*}

\begin{table*}
\setlength{\tabcolsep}{4pt}
\caption{Selected hyper-parameters for metric-based meta-learning algorithms in 20-shot setting.}
\label{table:graph}
\begin{center}
\begin{footnotesize}
\begin{sc}
\begin{tabular}{llccccc}
\toprule
\multicolumn{2}{c}{\textbf{method}} & \textbf{Task step} & \textbf{Adapt step} & \textbf{Task LR} & \textbf{Adapt LR} & \textbf{$|$Layers$|$} \\
\midrule
\multirow{3}{*}{\begin{turn}{90}Mol\end{turn}} 
& MatchNet   				 & 25  &  10 & 0.01 & 0.1 & 2  \\
& ProtoNet    				  & 50  &  50 & 0.01 & 0.01 & 3  \\
& RelationNet      			& 50  &  50 & 0.001 & 0.01 & 2  \\
\midrule
\multirow{3}{*}{\begin{turn}{90}Bio\end{turn}} 
& MatchNet   				 & 50  &  50 & 0.001 & 0.01 & 2  \\
& ProtoNet    				  & 50  &  10 & 0.001 & 0.1 & 2  \\
& RelationNet      			& 25  &  25 & 0.01 & 0.1 & 3  \\
\midrule
\multirow{3}{*}{\begin{turn}{90}Social\end{turn}} 
& MatchNet   				 & 50  & 25 & 0.01 & 0.1  & 3  \\
& ProtoNet    				  & 25  & 10 & 0.01 & 0.1  & 3  \\
& RelationNet      			& 50  & 10 & 0.01 & 0.01  & 3  \\
\bottomrule
\end{tabular}
\end{sc}
\end{footnotesize}
\end{center}
\end{table*}

\begin{table*}
\caption{Mean and Standard Deviation of meta-test accuracy on Molecules benchmark after ten runs.}
\label{table:graph}
\begin{center}
\begin{footnotesize}
\begin{sc}
\begin{tabular}{clccccc}
\toprule
\multicolumn{2}{c}{\textbf{method}} & \textbf{1-shot} & \textbf{5-shot} & \textbf{10-shot} &\textbf{ 20-shot} \\
\midrule
& Empirical Upper Bound  & \multicolumn{4}{c}{\textbf{67.62 $\pm$ 7.67}} \\
\midrule
\multirow{3}{*}{\begin{turn}{90}Sup\end{turn}}  
& GCN             & 56.68 $\pm$ 6.16 & 56.83 $\pm$ 6.03  & 56.59 $\pm$ 6.28 & 56.69 $\pm$ 6.06  \\
& GAT             & 56.48 $\pm$ 5.99 & 56.80 $\pm$ 5.87  & 56.66 $\pm$ 5.89 & 56.76 $\pm$ 6.02 \\
& GIN             & 57.12 $\pm$ 6.73 & 57.59 $\pm$ 6.61  & 57.77 $\pm$ 6.72 & 58.96 $\pm$ 4.96 \\
\midrule
\multirow{4}{*}{\begin{turn}{90}Uns\end{turn}}
& InfoGraph       & 57.11 $\pm$ 6.89 & 57.44 $\pm$ 6.61 & 57.28 $\pm$ 7.16 & 57.06 $\pm$ 6.66   \\ 
& MVGRL           & 58.89 $\pm$ 6.68 & 58.78 $\pm$ 5.33 & 58.83 $\pm$ 6.06 & 58.39 $\pm$ 6.33   \\
& GSFE            & 56.74 $\pm$ 5.93 & 56.93 $\pm$ 6.24 & 56.98 $\pm$ 6.85 & 57.21 $\pm$ 6.43   \\ 
& GCC             & 56.61 $\pm$ 5.88 & 56.84 $\pm$ 5.97 & 57.05 $\pm$ 6.32 & 57.36 $\pm$ 6.14   \\
\midrule
\multirow{9}{*}{\begin{turn}{90} Meta\end{turn}}
& MatchNet        & 58.89  $\pm$  6.31  & 59.33  $\pm$  5.71  & 59.50  $\pm$  6.83 &  60.06  $\pm$  6.87\\ 
& ProtoNet        & 58.72  $\pm$  6.75  & 59.11  $\pm$  6.80  &  59.28  $\pm$  4.89 &  59.44  $\pm$  6.13\\
& RelationNet     & 58.78 $\pm$   7.73  & 59.06 $\pm$   6.96  & 59.17 $\pm$   6.96 & 60.11 $\pm$   5.02	 \\
& MAML            & 56.00  $\pm$  6.71  & 56.78  $\pm$  6.76  &57.28  $\pm$  5.95  & 57.39  $\pm$  6.76 \\
& MetaSGD         & 56.89  $\pm$  5.79  & 57.11  $\pm$  6.68  & 57.39  $\pm$  6.91  & 58.33  $\pm$  4.89 \\
& MetaSpecGraph   & $-$  & 58.64  $\pm$  6.11  & 58.83  $\pm$  5.48	 & 60.59  $\pm$  5.89 \\
\cmidrule{2-6}
& MatchNet + Our Encoder   & \textbf{62.11 $\pm$ 6.14} & 62.28 $\pm$ 6.06 & 62.52 $\pm$ 6.03 & 62.87 $\pm$ 5.98  \\ 			
& ProtoNet  + Our Encoder    &  61.33 $\pm$ 6.91 & 61.67 $\pm$ 5.99 & 61.83 $\pm$ 5.14 & 62.44 $\pm$ 5.97  \\
& RelationNet  + Our Encoder      & \textbf{62.11 $\pm$ 5.05} & \textbf{62.50 $\pm$ 4.13} & \textbf{62.83 $\pm$ 6.82}  & \textbf{63.16 $\pm$ 5.16}  \\
& MAML + Our Encoder                    & 58.39 $\pm$ 0.68 & 59.00 $\pm$ 1.16 &  59.56 $\pm$ 1.40 & 60.22 $\pm$ 1.08 \\
& MetaSGD + Our Encoder                       &  58.67 $\pm$ 1.42 &  59.28 $\pm$ 1.27 & 59.39 $\pm$ 0.86 & 60.56 $\pm$ 1.20 \\
\bottomrule
\end{tabular}
\end{sc}
\end{footnotesize}
\end{center}
\end{table*}

\begin{table*}
\caption{Mean and Standard Deviation of meta-test accuracy on Molecules benchmark after ten runs.}
\label{table:graph}
\begin{center}
\begin{footnotesize}
\begin{sc}
\begin{tabular}{llccccc}
\toprule
\textbf{method} & \textbf{view} & \textbf{1-shot} & \textbf{5-shot} & \textbf{10-shot} &\textbf{ 20-shot} \\
\midrule
MatchNet                         & \textbf{X} & 58.89  $\pm$  6.31  & 59.33  $\pm$  5.71  & 59.50  $\pm$  6.83 &  60.06  $\pm$  6.87\\
ProtoNet                         & \textbf{X}  & 58.72  $\pm$  6.75  & 59.11  $\pm$  6.80  &  59.28  $\pm$  4.89 &  59.44  $\pm$  6.13\\
RelationNet                     & \textbf{X}  & 58.78 $\pm$   7.73  & 59.06 $\pm$   6.96  & 59.17 $\pm$   6.96 & 60.11 $\pm$   5.02	 \\
MAML                       		& \textbf{X} & 56.00  $\pm$  6.71  & 56.78  $\pm$  6.76  &57.28  $\pm$  5.95  & 57.39  $\pm$  6.76 \\
MetaSGD                    		& \textbf{X}  & 56.89  $\pm$  5.79  & 57.11  $\pm$  6.68  & 57.39  $\pm$  6.91  & 58.33  $\pm$  4.89 \\
\midrule
MatchNet 						& \textbf{X, U} & 60.15  $\pm$  6.35  & 60.37  $\pm$  6.70  & 60.73  $\pm$  5.62 &  61.42  $\pm$  6.25\\
ProtoNet                          & \textbf{X, U}  & 59.89  $\pm$  6.46  & 60.14  $\pm$  5.85  &  60.28  $\pm$  5.55 &  60.63  $\pm$  5.41\\
RelationNet                      & \textbf{X, U}  & 60.93 $\pm$   6.67  & 61.06 $\pm$   5.23  & 61.19 $\pm$   4.84 & 61.41 $\pm$   6.07	 \\
MAML 							& \textbf{X, U} & 57.12  $\pm$  5.89  & 57.48  $\pm$  6.43  & 57.98  $\pm$  5.89  & 58.12  $\pm$  6.34 \\
MetaSGD                        & \textbf{X, U}  & 57.64  $\pm$  6.56  & 57.72  $\pm$  6.47  & 58.26  $\pm$  5.97  & 58.65  $\pm$  6.67 \\
\midrule
MatchNet                       & \textbf{X, Z} & 61.47  $\pm$  7.82  & 60.68  $\pm$  6.02  & 61.11  $\pm$  5.81 &  61.76  $\pm$  6.12\\
ProtoNet                        & \textbf{X, Z}  & 60.72  $\pm$  6.65  & 60.94  $\pm$  7.35  &  61.28  $\pm$  5.78 &  61.64  $\pm$  4.42\\
RelationNet       	          & \textbf{X, Z}  & 61.88 $\pm$   5.74  & 61.97 $\pm$   7.75  & 52.02 $\pm$   7.82 & 62.11 $\pm$   5.21	 \\
MAML                      		 & \textbf{X, Z} & 57.85  $\pm$  6.31  & 58.13  $\pm$  6.48  &58.24  $\pm$  5.91  & 59.09  $\pm$  5.62 \\
MetaSGD                      & \textbf{X, Z}  & 57.91  $\pm$  7.11  & 58.11  $\pm$  7.66  & 58.35  $\pm$  6.88  & 58.92  $\pm$  5.48 \\
\midrule
MatchNet   				 & \textbf{X, U, Z}  & \textbf{62.17 $\pm$ 6.14} & 61.50 $\pm$ 6.06 & 61.94 $\pm$ 6.03 & 62.33 $\pm$ 5.98  \\
ProtoNet    				  & \textbf{X, U, Z}  &  61.33 $\pm$ 6.91 & 61.67 $\pm$ 5.99 & 61.83 $\pm$ 5.14 & 62.44 $\pm$ 5.97  \\
RelationNet      			 & \textbf{X, U, Z}  & 62.11 $\pm$ 5.05 & \textbf{62.50 $\pm$ 4.13} & \textbf{62.83 $\pm$ 6.82}  & \textbf{63.16 $\pm$ 5.16}  \\
MAML                     	 & \textbf{X, U, Z}  & 58.39 $\pm$ 6.68 & 59.00 $\pm$ 6.16 &  59.56 $\pm$ 6.40 & 60.22 $\pm$ 6.08 \\
MetaSGD                    & \textbf{X, U, Z} &  58.67 $\pm$  5.42 &  59.28 $\pm$  6.27 & 59.39 $\pm$  6.86 & 60.56 $\pm$  6.20 \\
\bottomrule
\end{tabular}
\end{sc}
\end{footnotesize}
\end{center}
\end{table*}

\begin{table*}
\caption{Mean and Standard Deviation of meta-test accuracy on Bioinformatics benchmark after ten runs.}
\label{table:graph}
\begin{center}
\begin{footnotesize}
\begin{sc}
\begin{tabular}{llccccc}
\toprule
\textbf{method} & \textbf{view} & \textbf{1-shot} & \textbf{5-shot} & \textbf{10-shot} &\textbf{ 20-shot} \\
\midrule
MatchNet                 & \textbf{X} &  54.83  $\pm$  7.66 &  55.62  $\pm$  7.60 & 55.92  $\pm$  6.67 & 56.04  $\pm$  7.78 \\
ProtoNet                  & \textbf{X} & 54.71  $\pm$  8.86  & 55.75  $\pm$  7.84  & 55.96  $\pm$  6.73 & 55.50  $\pm$  9.65 \\
RelationNet             & \textbf{X} & 54.93 $\pm$   8.55  & 55.92 $\pm$   8.69  & 56.02 $\pm$   7.69 & 56.15 $\pm$   7.81 \\
MAML                     & \textbf{X} & 53.83 $\pm$  9.62 & 54.46  $\pm$  6.77  & 54.50  $\pm$  8.77  & 54.79  $\pm$  8.90 \\
MetaSGD                 & \textbf{X} & 53.83  $\pm$  8.79 &  54.21  $\pm$  7.70 & 54.67  $\pm$  9.72 & 54.71  $\pm$  7.90 \\
\midrule
MatchNet                 & \textbf{X, U} &  56.72  $\pm$  8.11 &  56.44  $\pm$  8.39 & 56.89  $\pm$  6.39 & 57.06  $\pm$  8.50 \\
ProtoNet                  & \textbf{X, U} & 56.39  $\pm$  8.61  & 56.06  $\pm$  7.50  & 56.50  $\pm$  7.83 & 56.33  $\pm$  6.33 \\
RelationNet             & \textbf{X, U} & 56.83 $\pm$   8.11  & 56.94 $\pm$   9.83  & 56.50 $\pm$   7.56 & 56.94 $\pm$   8.33 \\
MAML                     & \textbf{X, U} & 55.62 $\pm$  8.08 & 55.17  $\pm$  8.50  & 55.87  $\pm$  9.83  & 55.92  $\pm$  7.83 \\
MetaSGD                 & \textbf{X, U} & 54.75  $\pm$  7.92 &  55.50  $\pm$  7.83 & 55.21  $\pm$  7.29 & 55.58  $\pm$  7.62 \\
\midrule
MatchNet                 & \textbf{X, Z} &  57.46  $\pm$  6.29 &  57.08  $\pm$  7.60 & 56.62  $\pm$  7.67 & 57.14  $\pm$  7.87 \\
ProtoNet                   & \textbf{X, Z} & 56.68  $\pm$  7.11  & 56.87  $\pm$  6.56  & 57.35 $\pm$  7.73 & 57.47  $\pm$  8.46 \\
RelationNet              & \textbf{X, Z} & 56.98 $\pm$   8.34  & 57.77 $\pm$   8.34 & 57.13 $\pm$   8.76 & 58.25 $\pm$   7.61 \\
MAML                      & \textbf{X, Z} & 55.47 $\pm$  7.67 & 55.79  $\pm$  7.67  & 55.40  $\pm$  8.71  & 56.72  $\pm$  7.87 \\
MetaSGD                  & \textbf{X, Z} & 54.63  $\pm$  9.21 &  55.46  $\pm$  8.50 & 56.47  $\pm$  8.79 & 56.77  $\pm$  8.96 \\
\midrule
MatchNet                 & \textbf{X, U, Z}  & \textbf{59.12 $\pm$ 7.00}  & 58.67 $\pm$ 9.77   & 57.71 $\pm$ 8.72  & 57.62 $\pm$ 6.97  \\
ProtoNet                   & \textbf{X, U, Z}  & 57.17 $\pm$ 7.76   & 57.58 $\pm$ 8.89   & 57.79 $\pm$ 8.76  & 58.17 $\pm$ 7.88  \\
RelationNet              & \textbf{X, U, Z}  & 58.83 $\pm$ 8.03  & \textbf{58.83 $\pm$ 9.68 }  & \textbf{59.29 $\pm$ 7.87}  & \textbf{59.82 $\pm$ 7.93}  \\
MAML                      & \textbf{X, U, Z}  & 56.00 $\pm$ 8.74   & 56.21 $\pm$ 7.76  & 56.37 $\pm$ 8.81   & 57.04 $\pm$ 7.85 \\
MetaSGD                  & \textbf{X, U, Z}  &  55.08 $\pm$ 8.67      & 56.12 $\pm$  8.23      & 57.08 $\pm$   8.86      &  57.33 $\pm$  8.86 \\
\bottomrule
\end{tabular}
\end{sc}
\end{footnotesize}
\end{center}
\end{table*}

\begin{table*}[h]
\caption{Mean and Standard Deviation of meta-test accuracy on Social Networks benchmark after ten runs.}
\label{table:graph}
\begin{center}
\begin{footnotesize}
\begin{sc}
\begin{tabular}{llccccc}
\toprule
\textbf{method} & \textbf{view} & \textbf{1-shot} & \textbf{5-shot} & \textbf{10-shot} &\textbf{ 20-shot} \\
\midrule
MatchNet                         & \textbf{X} &   62.25  $\pm$  9.14  & 62.42  $\pm$  9.98  & 62.92  $\pm$  9.22 &  63.33  $\pm$  9.92 \\
ProtoNet                          & \textbf{X} &   60.50  $\pm$  9.05  & 61.25  $\pm$  10.01  & 61.75  $\pm$  9.06 & 63.50  $\pm$  9.97 \\
RelationNet                      & \textbf{X} &   61.25 $\pm$   10.04  & 61.08 $\pm$   10.25 & 61.83 $\pm$   10.17 & 62.00 $\pm$   10.17 \\
MAML                            & \textbf{X} &   58.33  $\pm$  9.05  & 58.75  $\pm$  10.84  & 59.00  $\pm$  9.87 & 60.17  $\pm$  10.99 \\
MetaSGD                        & \textbf{X} &   59.25  $\pm$  10.15  & 59.33  $\pm$  9.50  & 59.83  $\pm$  9.89 & 60.25  $\pm$  9.88 \\
\midrule
MatchNet                         & \textbf{X, U} &   65.31  $\pm$  10.11  & 65.42  $\pm$  10.82  & 65.92  $\pm$  10.13 &  66.43  $\pm$  9.89 \\
ProtoNet                          & \textbf{X, U} &   65.14  $\pm$  10.21  & 65.37  $\pm$  9.12  & 65.88  $\pm$  9.14 & 66.11  $\pm$  10.95 \\
RelationNet                      & \textbf{X, U} &   65.44 $\pm$   9.26  & 65.74 $\pm$   9.18 & 65.91 $\pm$   9.18 & 66.66 $\pm$   10.83 \\
MAML                            & \textbf{X, U} &   61.14  $\pm$  9.17  & 61.68  $\pm$  10.74  & 62.41  $\pm$  10.74 & 63.68  $\pm$  10.41 \\
MetaSGD                        & \textbf{X, U} &   61.21  $\pm$  10.08  & 61.92  $\pm$  10.41  & 62.58  $\pm$  9.71 & 63.11  $\pm$  10.81 \\
\midrule
MatchNet                         & \textbf{X, Z} &   66.44  $\pm$  10.34  & 66.81  $\pm$  10.85  & 67.12  $\pm$  10.33 &  67.39  $\pm$  9.88 \\
ProtoNet                          & \textbf{X, Z} &   65.83  $\pm$  9.53  & 65.98  $\pm$  10.07  & 66.05  $\pm$  9.13 & 66.48  $\pm$  10.99 \\
RelationNet                      & \textbf{X, Z} &   66.02 $\pm$   10.31  & 66.29 $\pm$   9.31 & 66.81 $\pm$   10.21 & 67.33 $\pm$   9.12 \\
MAML                            & \textbf{X, Z} &   61.52  $\pm$  9.21  & 62.98  $\pm$  10.46  & 64.04  $\pm$  10.94 & 64.89  $\pm$  9.91 \\
MetaSGD                        & \textbf{X, Z} &   61.36  $\pm$  9.13  & 62.84  $\pm$  10.63  & 63.77  $\pm$ 10.83 & 63.96  $\pm$  9.89 \\
\midrule
MatchNet                        & \textbf{X, U, Z}  & \textbf{67.42 $\pm$ 10.37} & \textbf{67.50 $\pm$ 9.24} &  \textbf{68.08 $\pm$ 9.09} &  68.25 $\pm$ 9.18  \\
ProtoNet                          & \textbf{X, U, Z}  & 66.42 $\pm$ 10.21  & 66.92 $\pm$ 10.55  & 67.00 $\pm$ 9.13 & 67.50 $\pm$ 10.46 \\
RelationNet                     & \textbf{X, U, Z}  & 66.75 $\pm$ 9.89 & 67.08 $\pm$ 10.80 & 67.33 $\pm$ 10.33 &   \textbf{69.67 $\pm$ 10.05}  \\
MAML                 			 & \textbf{X, U, Z}  & 62.25 $\pm$ 10.12 & 64.75 $\pm$ 10.79 &  66.58 $\pm$ 10.93 & 66.83 $\pm$ 10.21 \\
MetaSGD                     	& \textbf{X, U, Z}  & 61.92$\pm$ 9.56  & 63.17 $\pm$ 10.51  & 64.42 $\pm$  10.16  & 64.83 $\pm$  10.06  \\
\bottomrule
\end{tabular}
\end{sc}
\end{footnotesize}
\end{center}
\end{table*}
\end{document}